\pdfoutput=1

\documentclass[11pt]{article}

\usepackage{ACL2023} 

\usepackage{times}
\usepackage{latexsym}

\usepackage[T1]{fontenc}

\usepackage[utf8]{inputenc}

\usepackage{microtype}

\usepackage{inconsolata}
\usepackage{hyperref}       
\usepackage{url}            
\usepackage{booktabs}       
\usepackage{amsfonts}       
\usepackage{nicefrac}       
\usepackage{xcolor}
\usepackage{epsfig}
\usepackage{algorithm}
\usepackage{algorithmic}
\usepackage{amsthm}
\usepackage{amsmath}
\usepackage{amssymb}
\usepackage{microtype}
\usepackage{mathtools}
\usepackage{float}
\usepackage{graphicx}
\usepackage{wrapfig}
\usepackage{makecell}
\usepackage{caption}
\usepackage{subcaption}
\usepackage{bm}
\usepackage{bbm}
\usepackage{multirow}

\newcommand{\ba}{$_\texttt{base}$}

\newcommand\tf[1]{\textbf{#1}}

\usepackage[noabbrev,capitalize]{cleveref} 

\crefname{equation}{equation}{equations}   
\crefname{footnote}{footnote}{footnotes}   
\crefname{line}{line}{lines}               
\crefname{section}{\S}{\S\S}
\Crefname{section}{\S}{\S\S}    
\crefformat{section}{#2\S#1#3}  
\Crefformat{section}{#2\S#1#3}
\crefrangeformat{section}{\S\S#3#1#4--#5#2#6}
\Crefrangeformat{section}{\S\S#3#1#4--#5#2#6}

 \title{Learning Multi-Step Reasoning by Solving Arithmetic Tasks}

\author{Tianduo Wang \and Wei Lu\\
  StatNLP Research Group\\
  Singapore University of Technology and Design \\
  \texttt{\{tianduo\_wang,luwei\}@sutd.edu.sg} \\}

\begin{document}
\maketitle
\begin{abstract}
  Mathematical reasoning is regarded as a necessary ability for Language Models (LMs).
  Recent works demonstrate large LMs' impressive performance in solving math problems.
  The success is attributed to their Chain-of-Thought (CoT) reasoning abilities, 
    i.e., the ability to decompose complex questions into step-by-step reasoning chains,
    but such ability seems only to emerge from models with abundant parameters.
  This work investigates how to incorporate relatively small LMs with the capabilities of multi-step reasoning.
  We propose to inject such abilities by 
  continually pre-training LMs on a synthetic dataset \tf{\textsc{MsAT}} which is composed of \tf{\underline{M}}ulti-\tf{\underline{s}}tep
  \tf{\underline{A}}rithmetic \tf{\underline{T}}asks.
  Our experiments on four math word problem datasets show the effectiveness of the proposed method in enhancing LMs' math reasoning abilities.\footnote{
    Our code and data are released at \url{https://github.com/TianduoWang/MsAT}.
    }
\end{abstract}

\section{Introduction}

  %

  Making Language Models (LMs) perform mathematical reasoning is a valuable, yet challenging research objective~\cite{hendrycks2021measuring,cobbe2021gsm8k}.
  Recently, we have witnessed large-scale LMs' impressive performance on a series of reasoning tasks via {\em chain-of-thought} prompting~\cite{wei2022cot}.
  This method elicits large LM's ability to decompose a complex problem into several intermediate steps.
  However, it is believed that such ability only emerges from sufficiently large models (empirically more than 100B parameters)~\cite{wei2022cot}.
  In this paper, 
    we examine how to incorporate moderate-sized LMs,
    e.g., RoBERTa~\cite{liu2019roberta}, with such multi-step reasoning ability via continual pre-training to improve the performance on math problems.

\begin{figure}[t]
    \captionsetup{type=figure}
    \centering
    \includegraphics[width=.87\linewidth]{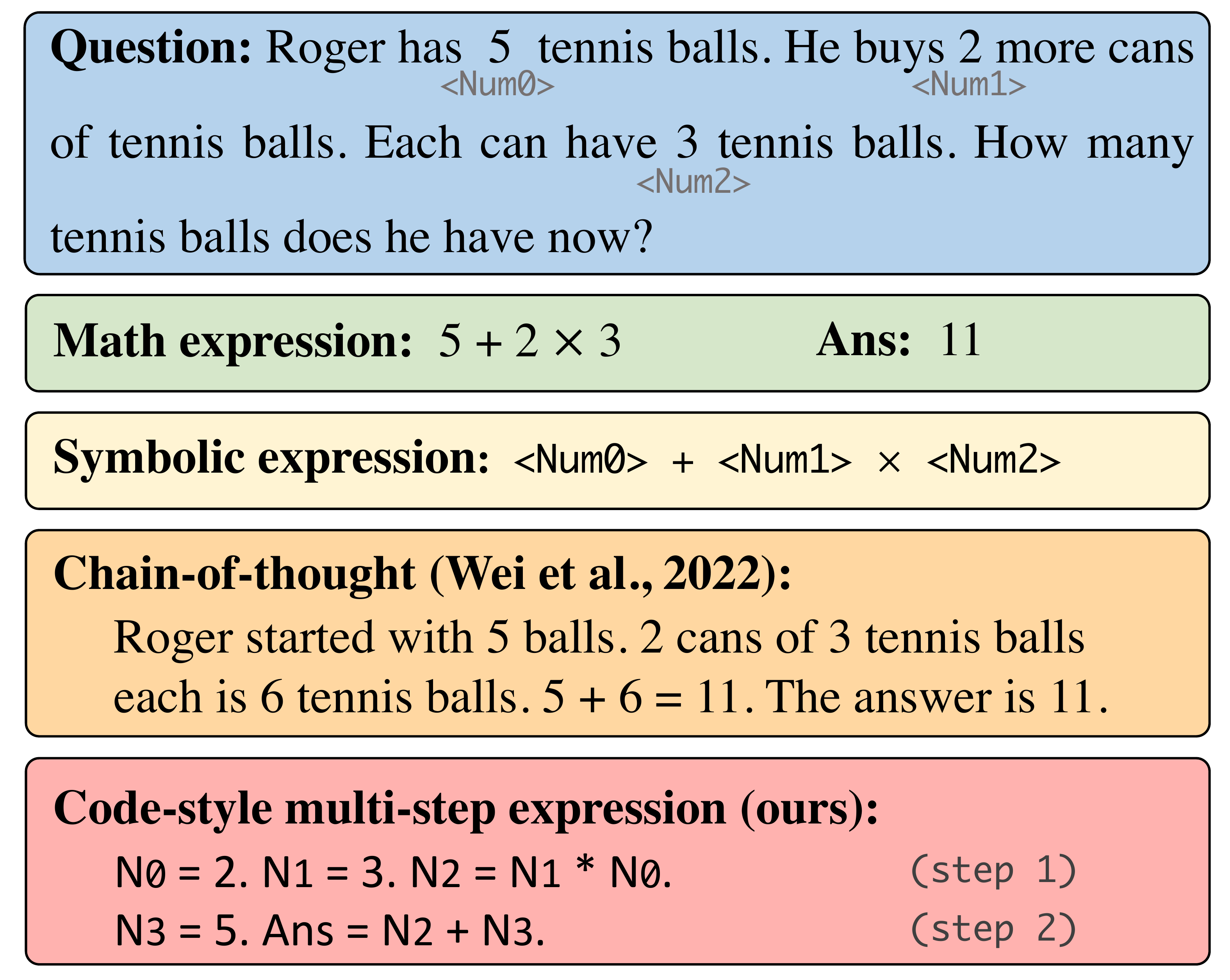}
    \caption{
        \label{fig:intro} 
        A math word problem example with different kinds of answers. 
        In \tf{Question}, \texttt{\small <Num0>}, \texttt{\small <Num1>}, and \texttt{\small <Num2>} are special tokens used for masking numbers.
    }
\end{figure}

  Correctly understanding numbers is a pre-requisite of mathematical reasoning abilities.
  But ~\citet{wallace2019nlp} shows that medium-sized LMs have a deficiency in numerical comprehension.
  To overcome this issue, previous works inject numerical reasoning skills into LMs following two approaches.
  The first is masking numbers with special tokens, 
    and generating symbolic expressions with a structured neural decoder~\cite{xie2019treedecoder,jie2022deductreason}.
  An example of such expression is provided in Figure~\ref{fig:intro}.
  The second strategy continually pre-trains LMs on synthetic numerical tasks,
    which requires models to learn how to perform computation involving numbers ~\cite{geva2020genbert,pi2022poet}.

  However, both approaches suffer from critical limitations.
  For symbolic methods, they neglect the information carried by the numbers,
    which could provide crucial hints for solving math problems~\cite{wu2021mathnum,liang2022mwpbert}.
  %
  %
  As for continual pre-training methods, 
    LMs' arithmetic skills are not reliable.
  Previous works indicate that such skills are highly influenced by the training data~\cite{razeghi2022impact} and hard for extrapolation~\cite{wallace2019nlp}.

  Motivated by these shortcomings, 
    we propose to first pre-train moderate-sized LMs on a synthetic dataset called \textsc{MsAT} (\underline{M}ulti-\underline{s}tep \underline{A}rithmetic \underline{T}asks)
    before downstream task fine-tuning.
  To make sure LMs capture the information carried by the numbers,
  we keep the numbers in the questions instead of masking them during both pre-training and fine-tuning.
  Instead of making LMs conduct computation internally, 
    \textsc{MsAT} encourages LMs to generate a series of intermediate steps
    leading to the answer.
  Experiments on four math word problem datasets with two backbone models
    demonstrate the effectiveness of our method in enhancing LMs' math reasoning performance.

\begin{figure}[t]
    \captionsetup{type=figure}
    \centering
    \includegraphics[width=.95\linewidth]{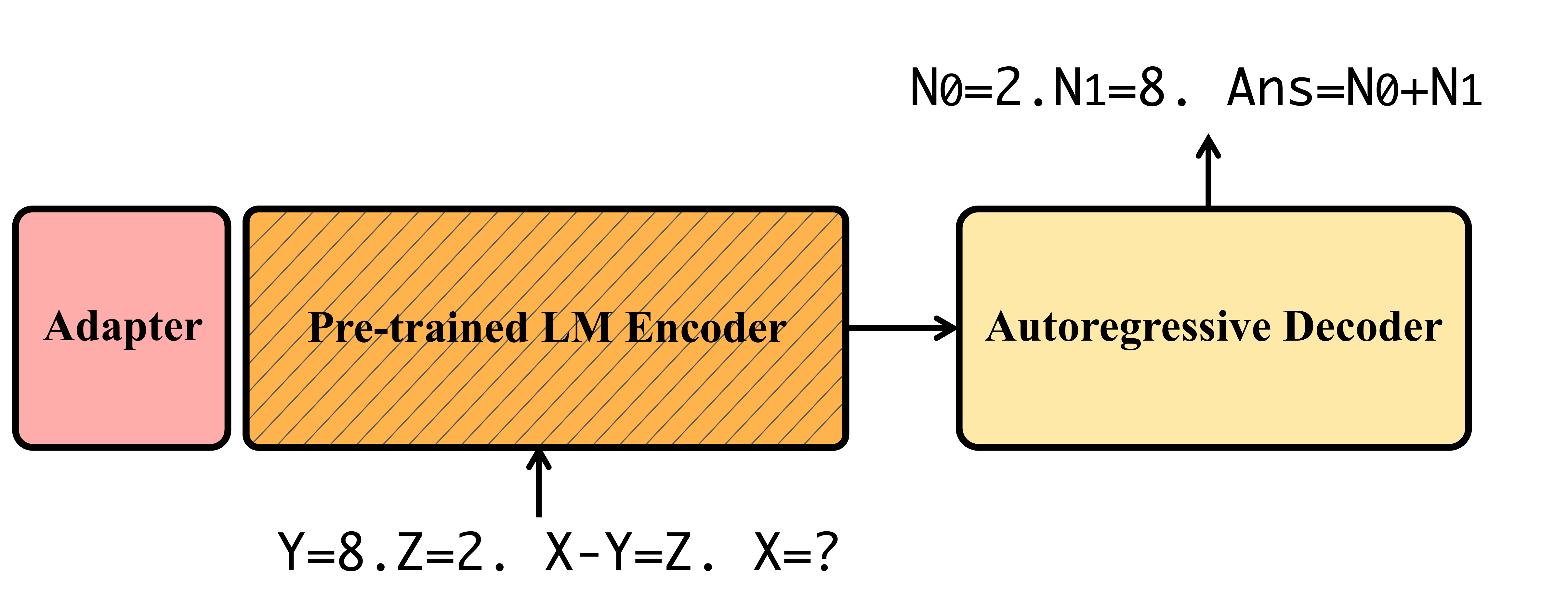}
    \vspace{-7pt}
    \caption{
        \label{fig:method} 
        An illustration of the continual pre-training process on our Seq2Seq model.
        We attach adapter modules to each layer of LM encoder and fix LM's parameters (shaded area) during pre-training.
        Tokens \texttt{\small N$_{\text{0}}$}, \texttt{\small N$_{\text{1}}$}, and \texttt{\small Ans} in the output are the variable names only used by the decoder.
        Our DAG structured model is similarly pre-trained with the only difference on the decoder part.
    }
    \vspace{-10pt}
\end{figure}

\section{Method}

  Our method essentially appends a continual pre-training stage before fine-tuning LMs on downstream tasks.
  The continual pre-training serves two purposes:
  first, we tokenize numbers digit-by-digit to improve LMs' numerical comprehension;
  second, we make LMs learn multi-step reasoning skills from the proposed synthetic task.

    \subsection{Digit tokenization for numbers}
      Sub-word tokenization methods, e.g.,  byte pair encoding (BPE)~\cite{sennrich2016bpe}, 
        is one of the reasons why moderated-sized LMs poorly understand numbers~\cite{wallace2019nlp}.
      BPE-based tokenizers split text based on the token frequency in the training corpus, which can be counter-intuitive when dealing with numbers.
      For example, numbers "\texttt{\small 520}" and "\texttt{\small 521}" will be tokenized into 
        ["\texttt{\small520}"] and ["\texttt{\small5}", "\texttt{\small21}"] respectively by the \texttt{RoBERTaTokenizer}\footnote{\url{https://huggingface.co/docs/transformers/model_doc/roberta}} of the \texttt{Transformers} library~\cite{wolf2020hf}.
      Such inconsistent tokenization strategy for numbers undermines LM's numerical understanding ability.
      Hence, we tokenize numbers digit-by-digit for both pre-training and fine-tuning.

    \subsection{Multi-step Arithmetic Tasks (\textsc{MsAT})}

      The core of our method is the synthetic task \textsc{MsAT} where LMs can learn multi-step reasoning skills.
      Like MWP tasks, \textsc{MsAT} can be formulated as a Seq2Seq task:
        the input of a \textsc{MsAT} example describes an arithmetic question,
        while the output is a reasoning chain leading to the answer.
      Specifically, each input sequence is composed of three components: {\em question context}, {\em equation}, and {\em question variable}.
      Equation is a sequence of symbols and operators ($+$, $-$, $\times$, $\div$, $=$) that builds equality relationship between symbols.
      Given an equation, only one of the symbols is set as the question variable,
        while other symbols will be listed in question context with their numerical values.

      The output sequence of \textsc{MsAT} is constructed in a code-style multi-step reasoning format.
      Each step consists of two sub-steps: {\em variable assignment} and {\em calculation}.
      In variable assignment, numbers appear in the input sequence are assigned to the variable names that are exclusive for decoder.
      In calculation, a new variable is generated from the calculation of the existing variables.
      This makes our outputs become executable Python code so that the numerical answer can be calculated by an external Python interpreter.
      Both inputs and outputs of \textsc{MsAT} are generated purely automatically.
      Details about the construction of \textsc{MsAT} are provided in~\cref{app:msat}.

\begin{table*}[t]
\small
\centering
\scalebox{0.95}{
\begin{tabular}{lllllllllll}
    \toprule
    \multirow{2}{*}{\textbf{Model}}  & \multicolumn{2}{l}{\textbf{MAWPS}} & \multicolumn{2}{l}{\textbf{ASDiv-A}} & \multicolumn{2}{l}{\textbf{SVAMP}} & \multicolumn{2}{l}{\textbf{SVAMP} (hard)}\\
    \cmidrule(lr){2-3}
    \cmidrule(lr){4-5}
    \cmidrule(lr){6-7}
    \cmidrule(lr){8-9}
    & Acc. & $\Delta$ & Acc. & $\Delta$ & Acc. & $\Delta$ & Acc. & $\Delta$ \\
    \toprule
    {\it Large language models}
    &\multicolumn{2}{l}{\tiny \texttt{(PaLM 540B)}} & \multicolumn{2}{l}{\tiny \texttt{(code-davici-002)}} & \multicolumn{2}{l}{\tiny \texttt{(PaLM 540B)}}     &  \\

    ~~~~~w/ Chain-of-Thought prompting
    & 93.3 & & 80.4 & & \tf{79.0} & & -  \\
    \midrule
    \multicolumn{5}{l}{\it{Seq2Seq models}}\\
    
    \textsc{RoBERTaGen}~\cite{lan2021mwptoolkit} & & & & \\
    
    ~~~~~~w/ symbolic masks
    & 88.4 & & 72.1 & & 30.3 & & 30.3$^{\heartsuit}$ \\
    
    ~~~~~~w/ digit tokenization
    & 84.1         & (\texttt{-}4.3) & 71.9          & (\texttt{-}0.2) & 27.6          & (\texttt{-}2.7) & 19.6          & (\texttt{-}10.7) \\
    
    \textsc{MsAT-RoBERTaGen (Ours)}
    & \tf{91.6} & (\texttt{+}3.2) & \tf{81.8} & (\texttt{+}9.7) & \tf{39.8} & (\texttt{+}9.5) & \tf{36.2} & (\texttt{+}5.9)\\

  \midrule
    \multicolumn{5}{l}{\it{DAG structured models}}\\

    \textsc{DeductReasoner}~\cite{jie2022deductreason}& & & & \\
      
    ~~~~~~w/ symbolic masks
    & 92.0 & & 85.0 & & 45.0 & & 45.0$^{\heartsuit}$ \\

    ~~~~~~w/ digit tokenization
    & 91.6 & (\texttt{-}0.4) & 84.1 & (\texttt{-}0.9) & 44.4 & (\texttt{-}0.6) & 42.8 & (\texttt{-}2.2)\\

    \textsc{MsAT-DeductReasoner (Ours)}
    & \tf{94.3} & (\texttt{+}2.3) & \tf{87.5} & (\texttt{+}2.5) & \tf{48.9} & (\texttt{+}3.9) & \tf{48.2} & (\texttt{+}3.2)\\
    \bottomrule
\end{tabular}}
\caption{
        \label{tab:main_results}
        Accuracy (\%) comparison between large language models (LLMs), backbone model baselines, and our method. 
        $\Delta$: performance gap compared with the symbolic mask baselines.
        ${\heartsuit}$: For baselines with symbolic masks, performance on SVAMP (hard) is the same as SVAMP because the actual numbers are replaced by symbolic tokens.
        The results of LLMs with chain-of-thought prompting are from~\citet{wei2022cot}.
    }
\end{table*}

    \subsection{Pre-training via adapter-tuning}

      Directly training on synthetic data that are largely different from the natural language corpus
        harms LMs' language prowess~\cite{geva2020genbert}.
      Therefore, we adopt a two-stage tuning strategy~\cite{wang2022diffaug} to inject reasoning skills into LMs.
      Specifically, we perform adapter-tuning~\cite{houlsby2019adapter} on \textsc{MsAT} and then jointly fine-tune adapter and LM backbone on downstream tasks.
      It mitigates catastrophic forgetting because LM's original parameters are largely preserved during adapter-tuning~\cite{houlsby2019adapter}.

      We consider two backbone models to verify the effectiveness of our method.
      In particular, we select a sequence-to-sequence (Seq2Seq) model~\cite{lan2021mwptoolkit} and a directed acyclic graph (DAG) structured model~\cite{jie2022deductreason} that both adopt RoBERTa\ba \ to encode the input questions.
      More details of these models are provided in~\cref{sec:exp_set}.
      Figure~\ref{fig:method} shows an overview of the proposed pre-training method.

\section{Experiments}
  Now we investigate whether our pre-training method facilitates models on Math Word Problem (MWP) solving tasks.
  All results are averaged over three different runs.

  \subsection{Experimental setup}\label{sec:exp_set}

    \paragraph{Existing datasets}
      
      We consider three commonly-used MWP datasets: 
        MAWPS~\cite{koncel2016mawps}, ASDiv-A~\cite{miao2020asdiv}, and SVAMP~\cite{patel2021svamp}. 
      The statistics of these datasets is provided in Table~\ref{tab:dataset_stat}.
      More details can be found in~\cref{app:exist}.
      We report five-fold cross-validation results for both MAWPS and ASDiv-A and test set accuracy for SVAMP 
        following previous practice~\cite{lan2021mwptoolkit,jie2022deductreason}.

\begin{table}[b]
    \begin{center}
    \centering
    \resizebox{\linewidth}{!}{
        \small
        \begin{tabular}{lccc}
        \toprule
           \multirow{2}{*}{\tf{Dataset}} &  \multirow{2}{*}{\tf{\# Data}} &  \tf{Avg. input} & \tf{Avg. output}\\
                                         &          &  \tf{length} &  \tf{reasoning steps}\\
        \midrule
            MAWPS         &   1,987&       30.3&        1.4\\
            ASDiv-A       &   1,217&       32.3&        1.2 \\
            SVAMP         &   1,000&       34.7&        1.2\\
        \bottomrule
        \end{tabular}
    }
    \caption{
        \label{tab:dataset_stat}
        \vspace{-10pt}
        Existing dataset statistics. 
    }
    \end{center}
\end{table}

    \paragraph{SVAMP (hard)}
      We find more than 85\% of the numbers in the above datasets are smaller than $10^2$.
      To investigate the extrapolation performance of the models trained with \textsc{MsAT},
        we create SVAMP (hard) from the original SVAMP dataset
        by replacing the numbers with much larger ones inspired by~\citet{gao2022pal}.
      More details about SVAMP (hard) and number distribution of the existing datasets are provided in~\cref{app:svamphard}.

    \paragraph{Models}
      We consider both sequence-to-sequence (Seq2Seq) models and directed acyclic graph (DAG) structured models as our backbone models.
      For Seq2Seq model, we choose \textsc{RoBERTaGen}~\cite{lan2021mwptoolkit}, 
        an encoder-decoder model with RoBERTa\ba \ as the encoder combined with a Transformer decoder.
      For DAG structured model, we choose \textsc{DeductReasoner}~\cite{jie2022deductreason} that combines RoBERTa\ba \ with a DAG decoder.
      In their original implementation, both models replace numbers with symbolic mask tokens.
      Hence, we additionally consider a baseline for each backbone model that uses actual numbers with digit tokenization.
      We name the models that are based on these two backbone models and pre-trained with our method as 
        \textsc{MsAT-RoBERTaGen} and \textsc{MsAT-DeductReasoner} respectively.
      We also compare our models to large LMs, e.g., PaLM~\cite{chowdhery2022palm} and Codex~\cite{chen2021codex}, with chain-of-thought prompting~\cite{wei2022cot}. 
      %
      All models are evaluated via greedy decoding.
      More implementation details, e.g., training hyperparameters, are provided in~\cref{app:implement}.

  \subsection{Main results}

    Table~\ref{tab:main_results} compares our models with backbone model baselines and large LMs.
    %
    %
    On all datasets, digit tokenization baselines consistently perform worse than their symbolic mask counterparts, 
      indicating the deficiency of the numeracy comprehension of the original RoBERTa model.
    However, the models trained with \textsc{MsAT} surpass both baselines by a large margin,
      which demonstrates the effectiveness of our pre-training method.

    \paragraph{SVAMP (hard)} 
      We can observe that, on SVAMP (hard), the accuracies of digital tokenization baselines decrease dramatically 
        (10.7 points drop for \textsc{RoBERTaGen} and 2.2 points drop for \textsc{DeductReasoner}) compared with baselines with symbolic masks, 
        while the models trained with \textsc{MsAT} still outperforms symbolic mask baselines by 5.9 and 3.2 points respectively. 
      This shows that not only does our models obtain better results than the baselines on the existing tasks, 
        but it is also more robust in handling out-of-distribution numbers.

    \paragraph{Compare with large language models}      
      We also observe that, on relatively simple tasks, i.e., MAWPS and ASDiv-A, 
        RoBERTa-based models can outperform large LMs.
      But for the more challenging task SVAMP, there is still a large performance gap.
      We believe this is because SVAMP requires models to have a better understanding of natural languages.
      \citet{jie2022deductreason} also reports that varying LM encoders results in significant performance disparities on SVAMP, 
        indicating that SVAMP performance is closely tied to model's natural language capabilities.

  \section{Pre-training analysis}\label{sec:ablation}

    In this section, we provide a careful analysis of our pre-training method from various perspectives to understand why it works.

\begin{figure}[t]
  \captionsetup{type=figure}
  \centering
  \includegraphics[width=.75\linewidth]{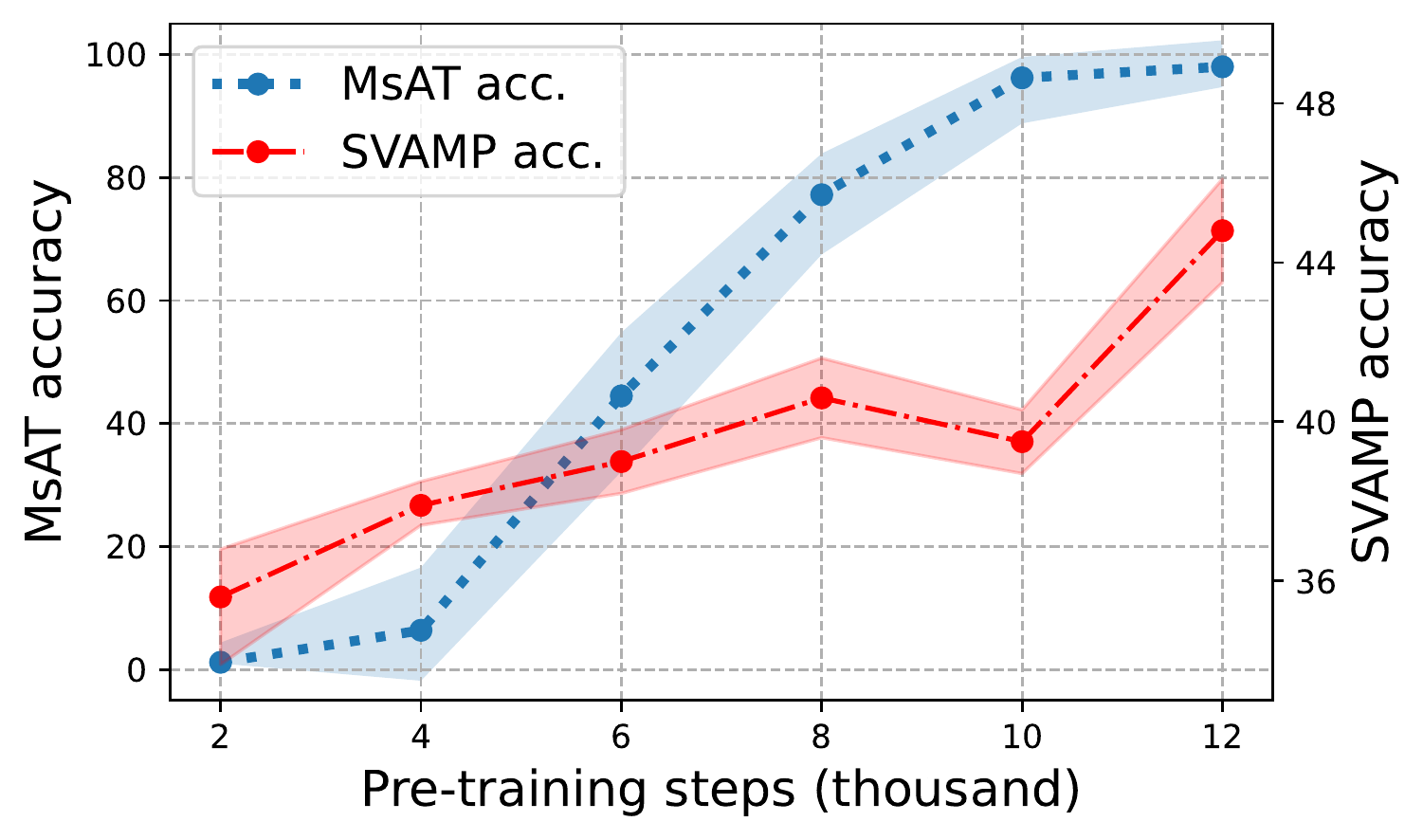}
  \caption{
      \label{fig:inter_perform} 
      Performance on \textsc{MsAT} and SVAMP with respect to the pre-training steps.
      Results are obtained from 3 different runs.
  }
\end{figure}

    \subsection{Pre-training task performance}\label{sec:pre_experiment}

      We visualize how the performance of pre-training task \textsc{MsAT} and one of the MWP tasks SVAMP changes with pre-training steps in Figure~\ref{fig:inter_perform}.
      It can be observed that the performance on both synthetic and natural language tasks tends to improve gradually as the number of pre-training steps increases.
      Figure~\ref{fig:inter_perform} demonstrates that LMs are capable of learning multi-step reasoning gradually from the synthetic task \textsc{MsAT}.
      The acquired multi-step reasoning ability can subsequently be transferred to the downstream MWP solving tasks, 
        enhancing performance during the fine-tuning phase.

\begin{figure}[t]
    \captionsetup{type=figure}
    \centering
    \includegraphics[width=.9\linewidth]{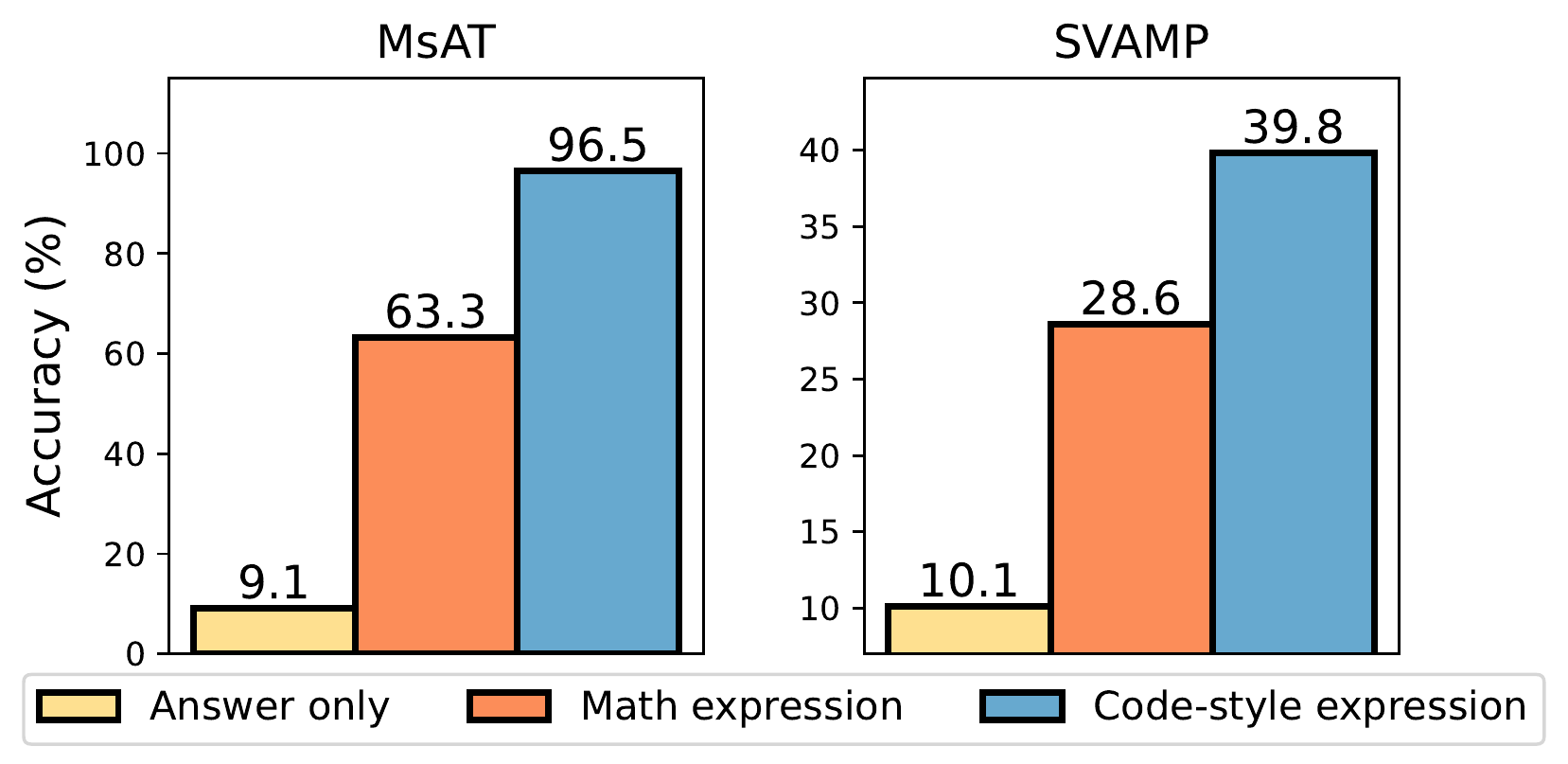}
    \caption{
        \label{fig:ablation}
        Comparison between different output expression formats.
        Results are obtained from our Seq2Seq model (with code-style expressions) and its variants.
    }
\end{figure}

  \subsection{Reasoning format of \textsc{MsAT}}

    The reasoning format of \textsc{MsAT} dictates the specific reasoning skills that LMs will acquire during pre-training.
    We demonstrate the superiority of our code-style multi-step reasoning format by comparing it with two different reasoning expressions.

    \paragraph{Effect of producing intermediate steps}
      While it is a common practice to train LMs 
        towards directly producing the numerical answers of the arithmetic questions~\cite{geva2020genbert,pi2022poet},
      a recent work shows that LMs' arithmetic skills are not reliable~\cite{razeghi2022impact}.
      To explore whether LMs can learn reasoning skills from \textsc{MsAT} without intermediate steps,
        we pre-train LMs on a variant of \textsc{MsAT} by replacing step-by-step output sequences with only numerical answers.
      Figure~\ref{fig:ablation} compares this model (answer only) with our model (code-style).
      Its poor performance on both \textsc{MsAT} and SVAMP confirms the necessity of producing intermediate reasoning steps during pre-training.

    \paragraph{Structured code-style expression}
      We next investigate the importance of applying the structured code-style reasoning expressions
        by comparing it with the less formatted math expressions.
      We argue that, compared with math expressions that only contain numbers and operators,
        our code-style expressions are more suitable for multi-step reasoning due to the structure information in the output sequences.
      Our experiments in Figure~\ref{fig:ablation} demonstrate the superiority of the code-style output expressions.
      We can see that 
        models with math expressions perform consistently worse than models with code-style multi-step reasoning format
        on both pre-training task \textsc{MsAT} and MWP solving task SVAMP.

\begin{figure}[t]
    \captionsetup{type=figure}
    \centering
    \includegraphics[width=.85\linewidth]{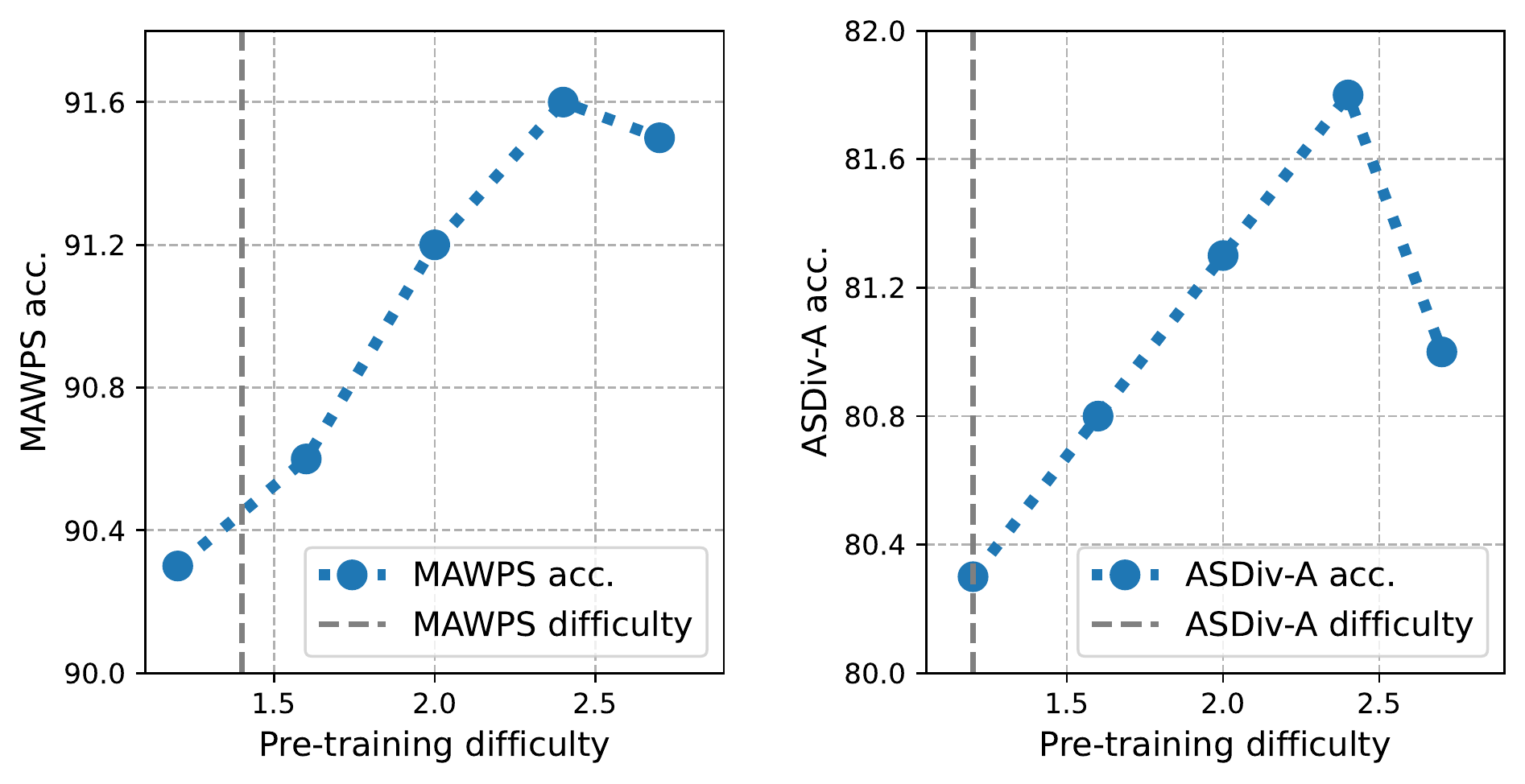}
    \vspace{-4pt}
    \caption{
        \label{fig:diff} 
        Performance on MAWPS and ASDiv-A with respect to pre-training difficulty.
        The difficulty levels of two MWP tasks are also added for reference.
    }
    \vspace{-5pt}
\end{figure}

    \subsection{Difficulty level of \textsc{MsAT}}

      %
      %
      Leveraging synthetic data for pre-training 
        provides the advantage of enabling highly customizable difficulty levels for the training data.
      Here we define the difficulty level of a reasoning task as the averaged reasoning steps that are required to solve the problems.
      From Figure~\ref{fig:diff}, we see that pre-training LMs on \textsc{MsAT}s that are harder than downstream tasks generally leads to better results.
      It's important to note that, broadly speaking, 
        the difficulty level of a reasoning task, 
        particularly those involving natural language, 
        is not solely determined by the number of reasoning steps.
      One example is that, though both ASDiv-A and SVAMP have an averaged reasoning steps of 1.2 (see Table~\ref{tab:dataset_stat}),
        SVAMP is considered more difficult as it requires high-level natural language understanding~\cite{patel2021svamp}.

    \subsection{Perform adapter-tuning on \textsc{MsAT}}

\begin{figure}[t]
    \captionsetup{type=figure}
    \centering
    \includegraphics[width=.95\linewidth]{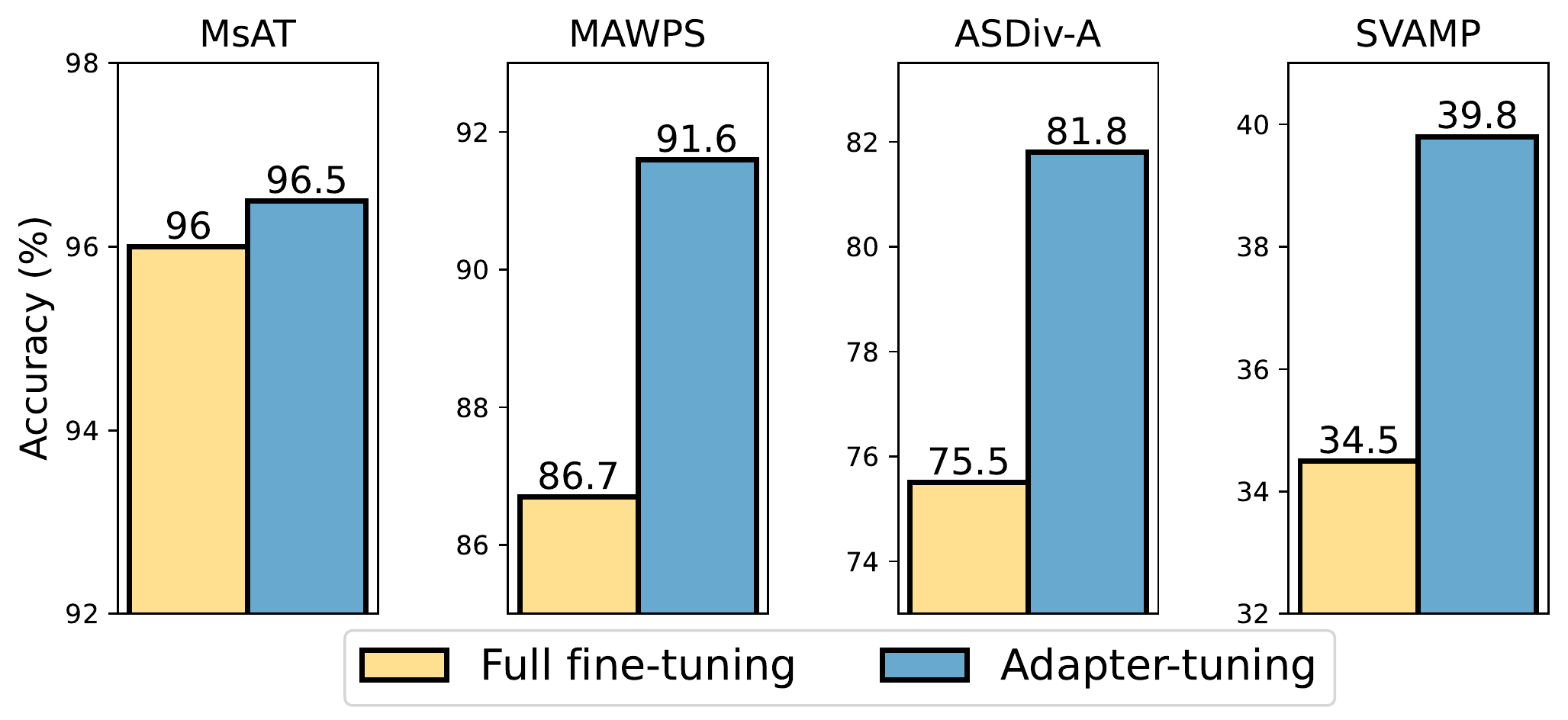}
    \vspace{-4pt}
    \caption{
        \label{fig:ft} 
        \textsc{MsAT} and downstream task performance comparison between full fine-tuning and adapter-tuning during pre-training.
    }
\end{figure}

      Tuning all parameters of LM encoders on synthetic data that are largely different from the pre-training corpus may lead to catastrophic forgetting~\cite{geva2020genbert}.
      To explore the importance of performing adapter-tuning on \textsc{MsAT}, 
        we create a variant of our method in which we perform full fine-tuning on \textsc{MsAT}.
      We compare this variant with our models in Figure~\ref{fig:ft}.
      It can be observed that both full fine-tuning and adapter-tuning can achieve good performance on \textsc{MsAT},
       but adapter-tuning outperforms fine-tuning on all downstream MWP datasets, 
        which demonstrates the benefits of performing adapter-tuning on \textsc{MsAT}.

\section{Related Work}

  In this work, we focus on improving moderate-sized LM's MWP performance by injecting multi-step reasoning ability.
  Hence, our work closely relates to both reasoning ability injection~\cite{geva2020genbert,pi2022poet}
    and MWP solving~\cite{xie2019treedecoder,patel2021svamp,jie2022deductreason}.

  \paragraph{Reasoning skills injection}
    %
    %
    This technique refers to continually pre-training LMs on certain intentionally-crafted tasks to enhance their reasoning abilities.
    GenBERT~\cite{geva2020genbert} pre-trains LMs on templated-based synthetic data to inject numerical skills into the LMs.
    PoET~\cite{pi2022poet} improves LMs' reasoning ability by pre-training them on tabular data towards imitating program executors.
    Both methods involve training LMs to produce numerical answers directly, which can be unreliable~\cite{razeghi2022impact}.
    Our work focuses on injecting into LMs the capability for solving complex arithmetic problems step-by-step.

  \vspace{+7pt}
  \paragraph{Solving MWP with specialized architectures}
    One of the research lines of MWP solving focuses on designing specialized achiectures 
      for math reasoning~\cite{xie2019treedecoder,lan2021mwptoolkit,jie2022deductreason}.
    For example, \citet{lan2021mwptoolkit} combines RoBERTa~\cite{liu2019roberta} with a Transformer~\cite{vaswani2017transformer} decoder,
      and \citet{jie2022deductreason} augments encoder-only LMs with a directed acyclic graph decoder.
    One of the shortages of such models is the information loss caused by masking actual numbers in the questions with symbolic tokens~\cite{wu2021mathnum}.
    In this work, we propose to represent actual numbers with digit tokenization, 
      and improve models' multi-step reasoning ability by pre-training them on a synthetic task \textsc{MsAT}.

\section{Conclusion}

  We propose a novel synthetic pre-training task, \textsc{MsAT},
    to incorporate LMs with multi-step reasoning skills that improve performance on MWP tasks.
  This pre-training task encourages LMs to generate intermediate reasoning steps instead of predicting final numerical answers directly.
  Our experiments show that the proposed method is effective in improving the moderate-sized LM's performance on MWP solving tasks.

\clearpage

\section*{Limitations} \label{limit}
  \paragraph{Limited number of operators considered}
    Following previous methods~\cite{lan2021mwptoolkit},
      we only consider binary operators ($+$, $-$, $\times$, and $\div$).
    As we adopt a code-style output format,
      it is possible to introduce other non-binary operators supported by the Python interpreter,
      e.g., \texttt{sum()} and \texttt{max()}.
    However, obtaining labeled data with such operators may require laborious efforts.
    We believe it is an interesting research question on exploring how to teach models to solve practical questions e.g., math word problems, by writing code in a low-resource setting~\cite{jie2023leveraging}.

  \paragraph{Limited performance due to greedy decoding}
    All the results we report in this work are produced via greedy decoding.
    A recent work~\cite{wang2022selfconsistency} reports that making large LMs generate multiple answers and selecting the answer with the most votes can boost performance by a large margin.
    However, performing beam search for symbolic neural reasoners, e.g., DeductReasoner, can be challenging
      in that searching space increases exponentially with the number of variables in the question~\cite{jie2022deductreason}.
    Designing effective beam search strategies for symbolic neural reasoners is a promising direction.


\section*{Acknowledgements}
We would like to thank the anonymous reviewers, our meta-reviewer, and senior area chairs for their insightful comments and support with this work.
We would also like to thank members of our StatNLP research group for helpful discussions.
This research/project is supported by the National Research Foundation Singapore and DSO National Laboratories under the AI Singapore Program (AISG Award No: AISG2-RP-2020-016), and Ministry of Education, Singapore, under its Academic Research Fund (AcRF) Tier 2 Programme (MOE AcRF Tier 2 Award No: MOE-T2EP20122-0011)

\bibliography{custom}
\bibliographystyle{acl_natbib}

\appendix

\section{Additional information about datasets}\label{app:datasets}
    In this section, we provide additional details about the datasets that we used in the experiments.
    %

\subsection{Construction of \textsc{MsAT}}\label{app:msat}

    The proposed \textsc{MsAT} is a synthetic Seq2Seq task where the inputs describe arithmetic questions
      and outputs are the solutions represented by a code-style multi-step reasoning format.
    Both inputs and outputs of \textsc{MsAT} can be generated automatically.
    To construct an example of \textsc{MsAT}, we first generate the input sequence and then produce the output solution accordingly.
    In all, we generate 85,000 examples and split them into 80,000 and 5,000 for training and evaluation respectively.

    \paragraph{Input sequence construction}
      We start by preparing a set of equation templates and each equation template contains no more than 3 binary operators ($+$, $-$, $\times$, and $\div$).
      By enumerating the possible combinations of operators, we obtain $4+4^2+4^3=84$ equation templates in total.
      The first step to construct an input arithmetic question is to instantiate an equation from an equation template.
      For example, given an equation template "{\small \texttt{<Num0> + <Num1> = <Num2>}}",
        we assign each variable a value that makes the equality hold and a variable name selected from the capitalized letters.
      The numbers in the questions are sampled from {\small \texttt{0}} to {\small \texttt{10,000}}.
      The last step is to randomly pick a variable as the question variable.
      Therefore, the resulting input arithmetic question may look like: "{\small \texttt{A=1. C=3. A+B=C. B?}}"

\begin{figure}[h]
    \vspace{+5pt}
    \captionsetup{type=figure}
    \centering
    \includegraphics[width=\linewidth]{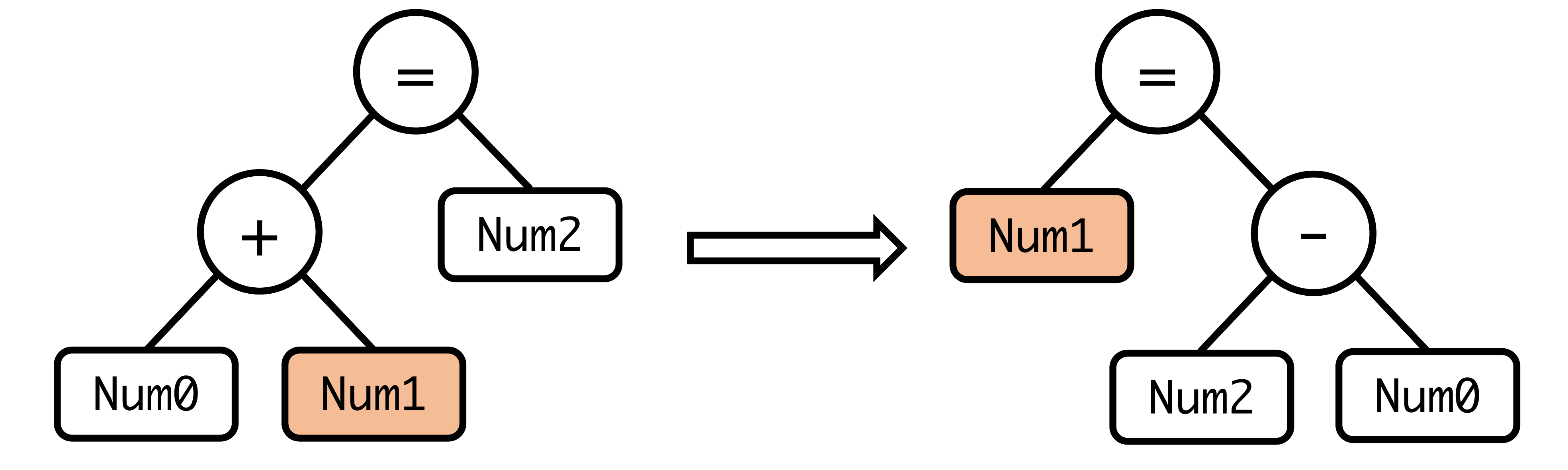}
    \vspace{-5pt}
    \caption{
        \label{fig:tree_inv}
        An illustration of the "tree inversion" algorithm that produces an output expression from an arithmetic question.
        The question variable is highlighted.
    }
\end{figure}

    \paragraph{Output sequence construction}
    Given an equation and a question variable, the output is first constructed as a math expression leading to the value of the question variable.
    Notice that an equation can be represented as a binary tree where the variables are the terminal nodes and operators are the non-terminal nodes.
    Hence, the output can be produced by a "tree inversion" algorithm (see Figure~\ref{fig:tree_inv}) from an equation and a question variable.
    %

\subsection{Existing datasets}\label{app:exist}

    \paragraph{MAWPS~\cite{koncel2016mawps}}
        It is a popular benchmark dataset for math word problems.
        We use the five-fold split provided by~\citet{lan2021mwptoolkit} for evaluation.

    \paragraph{ASDiv-A~\cite{miao2020asdiv}}
        This is an English math word problem task containing various linguistic patterns and problem categories.
        We obtain the data and five-fold split from~\citet{patel2021svamp}.
    
    \paragraph{SVAMP~\cite{patel2021svamp}}
        It is a challenge set created for MWP model robustness evaluation. 
        The examples in SVAMP are from ASDiv-A with deliberately designed variations.
        Such variations include: 
            changing questions, adding irrelevant information, etc.
        Following the evaluation protocol suggested by~\citet{patel2021svamp},
            we train our models over 3,138 training examples from a combination of MAWPS and ASDiv-A.
    
\begin{figure}[t]
    \captionsetup{type=figure}
    \centering
    \includegraphics[width=.75\linewidth]{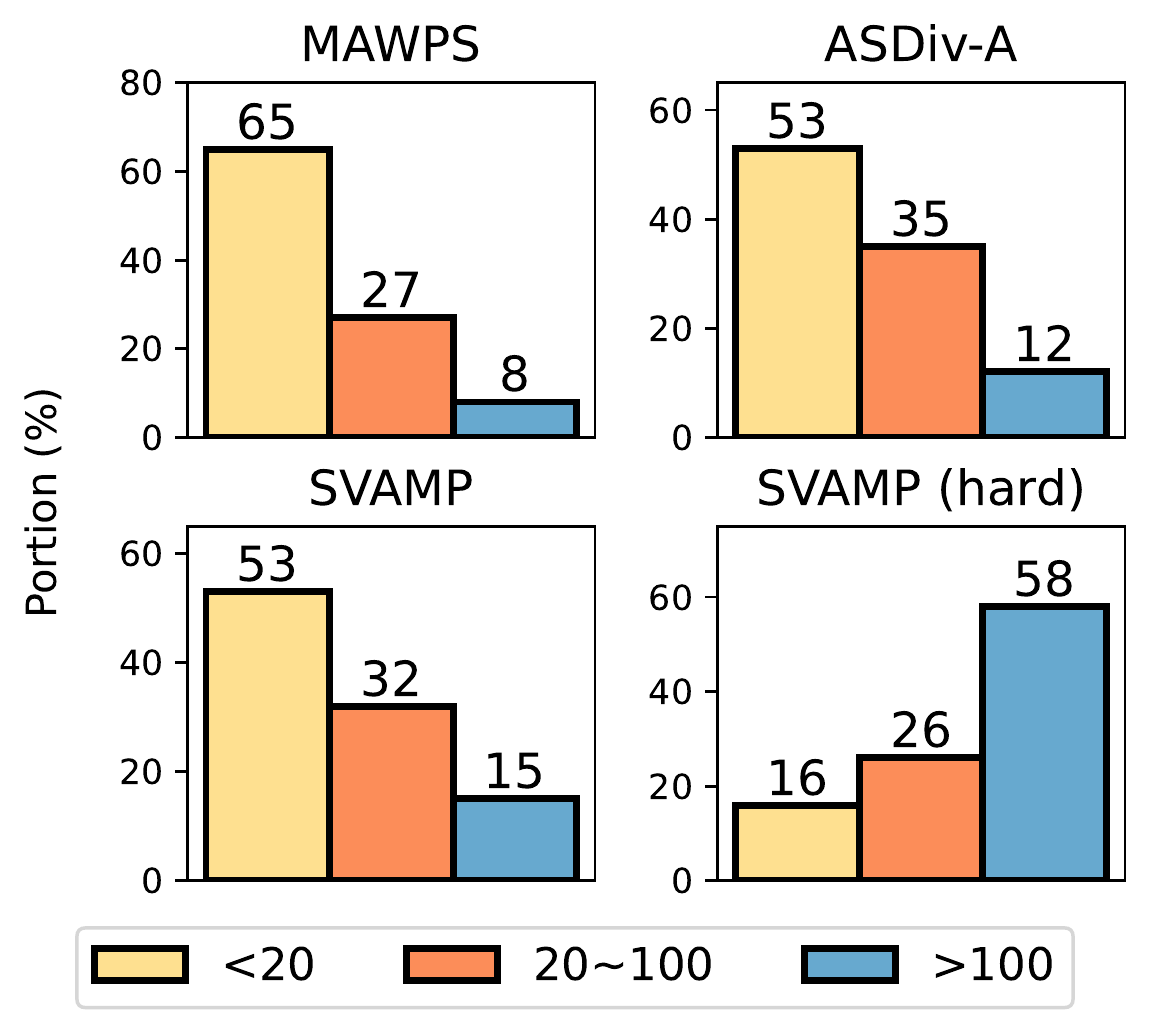}
    \vspace{-5pt}
    \caption{
        \label{fig:data_dist} 
        Number distribution for different datasets.
    }
    \vspace{-5pt}
\end{figure}

\subsection{SVAMP (hard)}\label{app:svamphard}
    SVAMP (hard) is used to evaluate models' extrapolation ability on the out-of-distribution numbers.
    We sample numbers from from {\small \texttt{10}} to {\small \texttt{10,000}}, 
        a significantly different range from the original one,
        to replace the original numbers in SVAMP.
    Every question in SVAMP (hard) corresponds to a question in SVAMP.
    Although it is straightforward to sample a large number and use it to replace the numbers,
        we expect the created questions to make sense.
    We achieve this by making sure the new numerical results have the same type as the original ones.
    For example, if the original numerical answer is a positive integer,
        then we make sure the new numerical answer is also a positive integer.
    We compare the number distribution of existing MWP datasets and SVAMP (hard) in Figure~\ref{fig:data_dist}.

\section{Implementation details}\label{app:implement}
    Our method is implemented in Python 3.8 with HuggingFace's Transformers~\cite{wolf2020hf} and PyTorch~\cite{paszke2019pytorch} libraries.
    %
    %
    All experiments can be conducted on one NVIDIA RTX 6000 GPU with 22 GB memory.

\subsection{Backbone Model implementation}

    For our \textsc{MsAT-RoBERTaGen} and \textsc{MsAT-DeductReasoner},
      we build the backbone models following the implementation provided by~\citet{lan2021mwptoolkit} and~\citet{jie2022deductreason} respectively.
    The encoders for both models are initialized with the pre-trained weights of RoBERTa\ba.
    The adapter modules~\cite{houlsby2019adapter} are added to each layer of the encoders with a bottleneck dimension of 64.
    More details about the mdoel architectures are provided in Table \ref{tab:model_hp}.

\begin{table}[h]
    \begin{center}
    \centering
    \resizebox{\linewidth}{!}{
        \begin{tabular}{lcc}
        \toprule
                             & \textsc{RoBERTaGen}  & \textsc{DeductReasoner} \\
        \midrule
            \# Params.       &   139.71 M           & 142.40 M \\
            \# Attention heads  &    8                 &    -                \\
            Hidden dim.      &    768               &    768              \\
            Feedforward dim. &    1024              &    768              \\
            \# Layers        &    2                 &    -                \\
            Activation       &    ReLU              &    ReLU             \\
            Dropout          &    0.1               &    0.1              \\
            Label smoothing  &    0.05              &    -                \\
            \# Constants     &    17                &    17               \\

        \bottomrule
        \end{tabular}
    }
    \caption{
        \label{tab:model_hp}
        Hyperparameters of model architectures.
    }
    \end{center}
\end{table}

\subsection{Training configurations}

\begin{table}[h]
    \begin{center}
    \centering
    \resizebox{\linewidth}{!}{
        \begin{tabular}{lcc}
        \toprule
                             & \textsc{Pre-training}  & \textsc{Fine-tuning} \\
        \midrule
            Batch size       &   32                 &            16          \\
            Max steps        &    10,000            &    50,000             \\
            Optimizer        &    \multicolumn{2}{c}{AdamW~\cite{loshchilov2017adamw}}                \\
            Weight decay     &   0.01               &       0.01        \\
            Max grad norm    &   0.1                &    1.0                 \\
            Learning rate    &   3e-5              &    1e-5                \\
            LR scheduler     &   Linear            &   Linear   \\     
        \bottomrule
        \end{tabular}
    }
    \caption{
        \label{tab:train_hp}
        Pre-training and fine-tuning hyperparameters.
    }
    \end{center}
\end{table}

\end{document}